\documentclass{article} % For LaTeX2e
\usepackage{iclr2026_conference,times}

\usepackage{amsmath,amsfonts,bm}

\def\eqref#1{equation~\ref{#1}}
\def\1{\bm{1}}

\DeclareMathAlphabet{\mathsfit}{\encodingdefault}{\sfdefault}{m}{sl}
\SetMathAlphabet{\mathsfit}{bold}{\encodingdefault}{\sfdefault}{bx}{n}

\DeclareMathOperator*{\argmin}{arg\,min}

\usepackage{hyperref}
\usepackage{url}
\hypersetup{
  hidelinks,
  pdftitle={J-CoT: Chain-of-Thought in J-Space},
  pdfauthor={Junde Wu, Jiayuan Zhu, Fengling Liu, Minhao Hu, and Jiazhen Pan}
}

\usepackage{booktabs}
\usepackage{multirow}
\usepackage{graphicx}
\usepackage{caption}
\usepackage{makecell}
\usepackage{array}
\usepackage{tabularx}
\usepackage{amsthm}
\usepackage{placeins}

\newtheorem{proposition}{Proposition}
\newtheorem{definition}{Definition}
\providecommand{\coloneqq}{\mathrel{\mathop:}=}

\title{J-CoT: Chain-of-Thought in J-Space}

\author{
\makebox[0.95\textwidth][c]{
Junde Wu\textsuperscript{1,2} \quad
Jiayuan Zhu\textsuperscript{1} \quad
Fengling Liu\textsuperscript{1} \quad
Minhao Hu\textsuperscript{1} \quad
Jiazhen Pan\textsuperscript{3}}
\\[4pt]
\makebox[0.95\textwidth][c]{
\textsuperscript{1}University of Oxford \quad
\textsuperscript{2}Imprint Lab \quad
\textsuperscript{3}Stanford University}
\\
\makebox[0.95\textwidth][c]{Correspondence: \texttt{jundewu@ieee.org}}
}

\iclrfinalcopy
\begin{document}

\maketitle

\begin{abstract}
Chain-of-thought prompting improves language-model reasoning by carrying
intermediate states across successive computation steps. However, relying on natural language as the only recurrent interface is overly restrictive, since many transient computations do not need to be fully verbalized.
Existing latent-reasoning methods remove this constraint by recurrently
propagating continuous hidden states. However, these methods pass a dense hidden
vector as a whole, without an explicit mechanism for selecting and organizing
the information needed by the next reasoning step. This motivates an
intermediate interface that remains linguistically grounded without requiring a
decoded sentence. We introduce
\textbf{J-CoT}, a recurrent reasoning framework built on \emph{J-space}, a
vocabulary-indexed coordinate system within the model's hidden
representations. Within each cycle, the model computes in its full hidden
space. At the cycle boundary, J-CoT expresses the intermediate state as
vocabulary-indexed coefficients, carries these coefficients forward as a
\emph{J-thought}, and maps them back into the model's hidden representation for
the next cycle.
J-CoT therefore requires neither a fluent intermediate rationale nor recurrence
over the complete hidden state. Under matched backbone and inference settings,
J-CoT-Zero matches or exceeds the strongest evaluated latent-reasoning baseline
on every benchmark, while J-CoT-Train obtains the highest score across the
evaluated mathematical, scientific, coding, and structured path-reasoning
tasks.
\end{abstract}

\section{Introduction}

Chain-of-thought (CoT) prompting has become one of the most effective
approaches for improving the reasoning capabilities of large language models
(LLMs)~\citep{wei2022chain,kojima2022large,wang2023selfconsistency}.
By generating intermediate reasoning steps before producing a final answer,
CoT substantially improves performance across a broad range of complex tasks
and has become a standard component of modern LLM inference.

Natural language, however, is an interface designed primarily for
communication rather than internal computation. When language is used as the
sole recurrent interface, every intermediate state must be serialized into a
coherent sequence of discrete tokens. The model must spend capacity on
grammatical form, discourse continuity, and explicit explanation, even when
these properties are unnecessary for the next reasoning step. More
fundamentally, it must commit partially formed, uncertain, or distributed
internal computations to a specific linguistic interpretation before they can
be carried forward. This requirement constrains the form of the state passed
between reasoning cycles. 
% We refer to this constraint as \emph{premature linguistic commitment}: an intermediate computation must take a complete linguistic form before its consequences for later computation are known.

Latent-reasoning methods address this limitation by removing the requirement
that every intermediate step be verbalized. Methods such as Coconut
recurrently propagate continuous hidden states between rounds of
computation~\citep{hao2024coconut}, allowing the model to preserve information
that may be difficult to express as natural language. However, the recurrent
state is a dense hidden vector passed as a whole. The interface has no explicit
mechanism for selecting and organizing the information needed by the next
reasoning step. 

%Dense hidden representations remain essential for computation
%\emph{within} a cycle; our question is whether the state communicated
%\emph{between} cycles can be more structured without being forced into decoded
%text.

Human reasoning offers an intuitive motivation for this intermediate design.
People do not formulate every internal transition as a complete,
well-formed sentence; but linguistically grounded partial hypotheses, salient concepts, and relations can guide subsequent thought before a final explanation is produced. It
suggests a useful design target to be a recurrent state that is linguistically
grounded but non-sentential, and that supports full hidden-space computation
after it is read back into the model.

Recent work on J-space provides a model-native candidate for such an
interface~\citep{gurnee2026workspace}. J-space is a vocabulary-indexed
coordinate system within the model's hidden representations. The vocabulary
index supplies a common address across layer-specific residual geometries,
while the coefficient state need not decode to a grammatical sequence. We
therefore view J-space as an intermediate interface between explicit linguistic
reasoning and dense latent recurrence, and introduce \textbf{J-CoT}, a recurrent
reasoning framework that operates through this interface.

\begin{figure}[t]
\centering
\includegraphics[width=\linewidth]{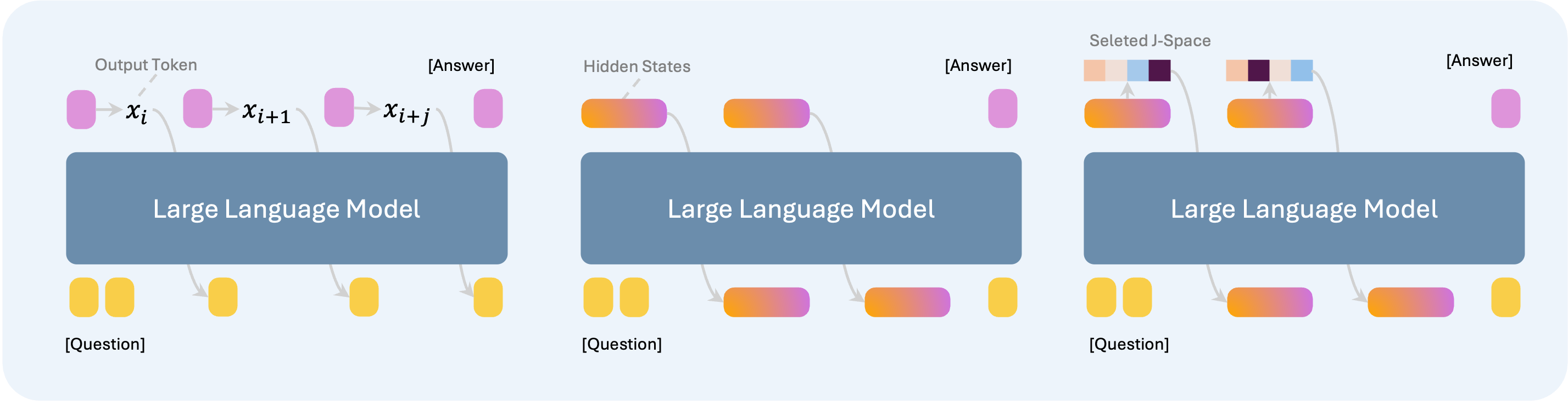}
\caption{Overview of recurrent reasoning interfaces. Explicit chain-of-thought
passes decoded tokens between steps, dense latent recurrence passes unrestricted
hidden states, and J-CoT passes a vocabulary-indexed coefficient state between
reasoning cycles; we call this state a J-thought.}
\label{fig:overview}
\end{figure}

At each reasoning cycle, the model first performs unrestricted computation in
its full hidden representation. J-CoT uses several non-decoded placeholder
positions, called \emph{carriers}, as the locations where the recurrent state is
read and written. At a selected write layer, J-CoT extracts coefficients over
the layer's J-space dictionary from the carrier activations. These coefficients
form the current \emph{J-thought}. To begin the next cycle, J-CoT reconstructs
the same coefficient state through the read-layer dictionary and introduces
the resulting carrier states into the Transformer. The recurrent state is
therefore neither a decoded rationale nor the complete hidden activation. It is
a vocabulary-indexed boundary state whose residual realization can change with
the layer at which it is read. Figure~\ref{fig:overview} contrasts this J-thought interface with
explicit chain-of-thought and dense latent recurrence.

J-CoT has both training-free and trainable configurations.
\emph{J-CoT-Zero} uses fixed interface components and requires no
J-CoT-specific optimization, whereas \emph{J-CoT-Train} optimizes the carrier
embeddings and a learned read gate while keeping the underlying Transformer
fixed. Unless explicitly stated otherwise, \emph{J-CoT} refers to the
training-free J-CoT-Zero configuration. We use \emph{J-CoT-Train} only when
referring to the variant with optimized carrier embeddings and a learned read
gate.

We evaluate J-CoT across mathematical reasoning, scientific question
answering, code generation, and structured path reasoning. Under matched
backbone and inference settings, J-CoT-Zero matches or exceeds the strongest
evaluated latent-reasoning baseline on every benchmark, and J-CoT-Train
improves further. 

\begin{itemize}
    \item We introduce \textbf{J-CoT}, a recurrent reasoning framework that
    carries a vocabulary-indexed \emph{J-thought} state across reasoning cycles.

    \item We develop a model-native read--write mechanism that carries this
    J-thought without decoded rationales or full hidden-state recurrence.

    \item We compare J-CoT with linguistic and latent-reasoning methods on
    mathematical, scientific, coding, and structured path-reasoning tasks.
\end{itemize}

\section{Related Work}

\paragraph{Explicit and search-based reasoning.}
Chain-of-thought (CoT) prompting elicits intermediate natural-language steps
before an answer, while zero-shot CoT obtains similar behavior through a fixed
instruction~\citep{wei2022chain,kojima2022large}. Subsequent work improves this
linguistic interface through sampling and aggregation, as in
self-consistency~\citep{wang2023selfconsistency}, or through explicit planning,
as in Plan-and-Solve prompting and Tree of
Thoughts~\citep{wang2023plansolve,yao2023tree}. These approaches differ in how
they generate, organize, or select candidate reasoning traces, but the
intermediate state passed to later computation remains a sequence of decoded
tokens. Theoretical analyses further connect CoT to an increase in the
effective sequential depth available to a Transformer~\citep{feng2023towards}.

\paragraph{Computation without a complete verbal rationale.}
Several lines of work give a language model additional computation without
requiring a conventional step-by-step explanation. Pause-token training inserts
learned special tokens before prediction~\citep{goyal2023think}, and filler-token
studies show that otherwise uninformative positions can provide useful
computational depth in suitable settings~\citep{pfau2024let}. Implicit-CoT
training progressively removes explicit reasoning steps while distilling their
computational role into the model~\citep{deng2024explicit}. At the architectural
level, looped Transformers repeatedly apply shared blocks and establish that
iteration itself can support nontrivial computation~\citep{giannou2023looped}.
These results motivate recurrent computation, but they do not specify the
vocabulary-indexed read--write interface studied here. J-CoT instead preserves
ordinary dense Transformer computation within each cycle and constrains only
the state communicated from one cycle to the next.

\paragraph{Continuous latent reasoning.}
Continuous-reasoning methods replace some or all intermediate language tokens
with hidden activations. Coconut feeds the final hidden state of one latent step
back as the input embedding of the next, and trains this recurrence with a
multi-stage curriculum~\citep{hao2024coconut}. CODI compresses explicit CoT into
a continuous latent process through self-distillation~\citep{shen2025codi},
whereas SIM-CoT adds step-level supervision to implicit reasoning states. Theory suggests that continuous states can
represent multiple candidate reasoning paths in superposition rather than
committing immediately to one verbal trace~\citep{zhu2025reasoning}. 

\paragraph{Interpreting and reusing internal representations.}
Representation-reading methods map intermediate activations to model outputs or
human-readable coordinates. The tuned lens learns layer-specific translators
that expose intermediate token predictions~\citep{belrose2023tuned}. J-space
identifies vocabulary-indexed directions shared across model components and
layers~\citep{gurnee2026workspace}. J-CoT uses this coordinate system as a
recurrent computational interface. This use of
J-space distinguishes J-CoT from both decoded-token recurrence and direct
dense-state recurrence.

\section{J-CoT: Recurrent Reasoning with J-Thoughts}
\label{sec:method}

J-CoT is motivated by the choice of interface between repeated computations.
Explicit chain-of-thought uses decoded text for this interface, while dense latent recurrence removes this commitment by carrying hidden activations.
J-CoT studies a third choice, it carries a vocabulary-indexed coefficient state
that can be realized in the residual geometry of different layers. We call this
state a \emph{J-thought}.

The design separates recurrent storage from within-cycle computation. Only the
state crossing the cycle boundary is represented as a J-thought. Once read back
into the model, it is processed by ordinary Transformer blocks in the full
hidden space. The method therefore does not assume that all useful computation
is expressible by the recurrent coefficient state; it assumes only that a component
useful to the next cycle can be extracted at the boundary. This section first
formalizes the recurrent-interface problem and its transport precondition, then
constructs the shared J-space coordinates, and finally describes how J-CoT
reads, revises, and uses a sequence of J-thoughts.

\subsection{Problem Formulation and Method Overview}

\paragraph{Motivation.}
J-space is a plausible recurrent interface because it provides a common index
for residual directions that otherwise change across layers. A vocabulary entry
$v$ is associated with a different residual direction $d_{\ell,v}$ at each
layer $\ell$, but the index $v$ is shared. If an activation can be expressed by
coefficients over these directions, the coefficients can serve as a common
address: they may be extracted in the geometry of one layer and reconstructed
in the geometry of another. This is the simple intuition behind J-CoT. The
method does not require the coordinate labels to be complete semantic concepts;
it requires only that the coefficient state be recoverable across the layers
used by the recurrence. A therotical guarantee for this transport property is given below.

\paragraph{Recurrent-interface problem.}
Let $M$ denote the number of recurrent carrier positions, and let
$F_{\theta,x}^{\ell_r:\ell_w}$ denote the Transformer computation between
a read layer $\ell_r$ and a later write layer $\ell_w$, conditioned on input
$x$. We seek a recurrent state space $\mathcal A$, a read map
$\mathsf R_{\ell_r}:\mathcal A\rightarrow\mathbb R^{M\times d}$, and a write
map $\mathsf W_{\ell_w}:\mathbb R^{M\times d}\rightarrow\mathcal A$. Their
composition defines

\begin{equation}
A_t
=
\mathcal T_{x,\phi}(A_{t-1}),
\qquad
\mathcal T_{x,\phi}
=
\mathsf W_{\ell_w}
\circ F_{\theta,x}^{\ell_r:\ell_w}
\circ\mathsf R_{\ell_r,\phi},
\qquad
A_0=0.
\label{eq:abstract-recurrence}
\end{equation}

Here $\phi$ denotes optional parameters of the recurrent interface; the
Transformer parameters $\theta$ are unchanged by this definition.

The modeling problem is to instantiate $\mathcal A$, $\mathsf R$, and
$\mathsf W$ so that a state written in the residual geometry of $\ell_w$ can be
read in the geometry of $\ell_r$ without first being decoded into language.
J-CoT chooses $\mathcal A$ to be a matrix of J-space coefficients. The read map
reconstructs those coefficients through the read-layer dictionary, and the
write map extracts new coefficients from carrier activations at the write
layer. Dense Transformer computation remains inside
$F_{\theta,x}^{\ell_r:\ell_w}$; only the recurrent boundary is represented in
J-space.

\paragraph{Transport precondition.}
Suppose each participating layer has a dictionary $D_\ell$, an extraction map
$\Phi_\ell$, and a reconstruction map $\Psi_\ell(a)=D_\ell a$. Their concrete
construction is given in Section~\ref{sec:canonical-jspace}. For a coefficient
vector $a$, define the recovery defect

\begin{equation}
\varepsilon_{\ell}(a)
\coloneqq
\|\Phi_{\ell}(\Psi_{\ell}(a))-a\|_2.
\label{eq:coordinate-recovery-defect}
\end{equation}

Small recovery defect is the mathematical precondition for using the same
coefficient identity across layers. The next proposition makes this condition
explicit; it is a conditional transport guarantee.

\begin{proposition}[Multi-hop stability of J-thought transport]
\label{prop:j-transport-stability}
Let $h_{\ell_0}=D_{\ell_0}a+\xi_{\ell_0}$ and
$\widehat a_0=\Phi_{\ell_0}(h_{\ell_0})$. Along any layer path
$p=(\ell_0,\ell_1,\ldots,\ell_K)$, recursively set
$\widehat a_i=\Phi_{\ell_i}(\Psi_{\ell_i}(\widehat a_{i-1}))$ for
$1\leq i<K$, and reconstruct
$T_p(h_{\ell_0})=\Psi_{\ell_K}(\widehat a_{K-1})$. Then
\begin{equation}
\|T_p(h_{\ell_0})-D_{\ell_K}a\|_2
\leq
\|D_{\ell_K}\|_2
\left(
\|\widehat a_0-a\|_2+
\sum_{i=1}^{K-1}\varepsilon_{\ell_i}(\widehat a_{i-1})
\right).
\label{eq:transport-fidelity}
\end{equation}
Consequently, sequential and direct transport differ by at most
$\|D_{\ell_K}\|_2\sum_{i=1}^{K-1}
\varepsilon_{\ell_i}(\widehat a_{i-1})$; exact recovery at the intermediate
layers gives path-independent transport.
\end{proposition}

The proof is given in Appendix. The bound identifies
what the subsequent construction must provide: an initial coefficient
extraction with limited error and dictionaries for which reconstruction followed
by re-extraction does not substantially alter the state. It isolates the
condition under which the extracted component can be reused at another layer, which motivates the J-CoT recurrent design.

\paragraph{Method overview.}
The remainder of the section instantiates the objects in
Eq.~\ref{eq:abstract-recurrence}. Section~\ref{sec:canonical-jspace} constructs
layer-specific J-lens dictionaries with a shared vocabulary index and defines
the extraction and reconstruction maps; this is what makes the write-layer
state readable at the earlier read layer. Section~\ref{sec:jthought-state} groups the resulting
coefficients across non-linguistic carrier positions and formally defines a
J-thought. Section~\ref{sec:jcot-recurrence} gives the complete
read--compute--write update and distinguishes J-CoT-Zero from J-CoT-Train.

\subsection{Canonical J-Space Construction}
\label{sec:canonical-jspace}

Let $h_{\ell,t}\in\mathbb R^d$ be the residual activation at layer $\ell$ and
position $t$, let $L$ be the final Transformer layer, and let
$\widetilde h_{L,t'}$ denote the final-normalized residual state supplied to the
unembedding matrix $W_U\in\mathbb R^{|\mathcal V|\times d}$. To identify how a
change at layer $\ell$ affects later output-facing representations, we estimate
the averaged downstream Jacobian

\begin{equation}
J_{\ell}
=
\mathbb E_{x,t,t'\geq t}
\left[
\frac{\partial \widetilde h_{L,t'}}{\partial h_{\ell,t}}
\right]
\in\mathbb R^{d\times d}.
\label{eq:averaged-jacobian}
\end{equation}

The columns of $W_U^{\top}$ are final-layer directions associated with
vocabulary entries. Pulling these directions back through $J_\ell$ gives the
corresponding directions at layer $\ell$. We form and normalize the dictionary

\begin{equation}
\widetilde D_{\ell}=J_{\ell}^{\top}W_U^{\top},
\qquad
d_{\ell,v}
=
\frac{\widetilde d_{\ell,v}}
{\|\widetilde d_{\ell,v}\|_2+\epsilon},
\qquad
D_{\ell}=[d_{\ell,1},\ldots,d_{\ell,|\mathcal V|}].
\label{eq:j-dictionary-construction}
\end{equation}

Thus, dictionary $D_\ell$ is specific to the residual geometry of layer
$\ell$, while its columns retain the vocabulary indexing induced by $W_U$. The
$\epsilon=10^{-8}$ term prevents division by a vanishing column norm. The
dictionaries are estimated once for each backbone on an unlabeled calibration
corpus and remain fixed during J-CoT training and evaluation. This fixed
construction is important for J-CoT-Zero in that cross-layer transport does not depend
on learning an additional alignment network from the downstream task data.
Appendix specifies the randomized Jacobian estimator,
calibration sampling, and normalization procedure.

Given an activation $h$ at layer $\ell$, extraction uses a nonnegative
elastic-net decomposition~\citep{zou2005regularization}; extraction and
reconstruction are defined by

\begin{equation}
\begin{aligned}
\Phi_{\ell}(h)
&\coloneqq
\argmin_{a\geq 0}
\frac{1}{2}\|h-D_{\ell}a\|_2^2
+\lambda_1\|a\|_1
+\frac{\lambda_2}{2}\|a\|_2^2,
\\
\Psi_{\ell}(a)
&\coloneqq D_{\ell}a.
\end{aligned}
\label{eq:j-coordinate-system}
\end{equation}

The map $\Phi_\ell$ identifies a nonnegative coefficient vector whose
layer-specific reconstruction approximates the activation. Nonnegativity makes
coefficient magnitude a consistent measure of positive coordinate presence,
rather than allowing the same contribution to be represented through
cancellation of positive and negative atoms. The $\ell_1$ penalty controls how
many coordinates are retained at the recurrent boundary, while the quadratic
term stabilizes the solution when dictionary directions are correlated. The
number of active coordinates is not prescribed, both support and coefficient
magnitudes are determined by the carrier activation in the current cycle.
Appendices~\ref{app:elastic-net-solver} and~\ref{app:elastic-net-gradient}
provide the numerical solver and differentiation details.

The reconstruction map $\Psi_\ell$ gives coefficients their residual
realization at a selected layer. Consequently, a state extracted at layer
$\ell$ can be transported to layer $m$ through
$\tau_{\ell\rightarrow m}=\Psi_m\circ\Phi_\ell$. The use of a shared index set
means J-CoT does not equate the residual vectors at the two
layers, but uses their dictionaries to express the same coefficient identity in
two different residual geometries.

\subsection{J-Thought State and Carrier Organization}
\label{sec:jthought-state}

\begin{definition}[J-thought]
For $M$ carrier positions, the J-thought after cycle $t$ is
$A_t\in\mathbb{R}_{\geq0}^{M\times|\mathcal V|}$. Row $A_{t,m:}$ contains the
J-space coefficients extracted from carrier $m$. The sequence
$(A_1,\ldots,A_{\tau(x)})$ is the \emph{J-thought trajectory} for input $x$.
\end{definition}

The vocabulary strings associated with active columns can be used to label a
J-thought, but the matrix is neither a token sequence nor a decoded rationale.
Rows are parallel coefficient channels, and columns are shared J-space
coordinates. No sequential linguistic meaning is assigned to the column order.
This terminology prevents the recurrent object from being conflated with a
generic memory construct or with an ordinary sequence of hidden tokens.
In practice, J-CoT appends
$M$ non-linguistic carrier positions after the input prompt. Carriers attend to
the complete prompt and to one another, and they participate in the backbone's
ordinary self-attention and feed-forward computation. They are excluded from
language-modeling targets and are never decoded. Their role is to expose a
small, consistent collection of residual activations from which the J-thought
can be written and into which it can be read.

Using multiple carriers avoids forcing the entire recurrent boundary through a
single activation. Different carriers can retain different coefficient supports
and can exchange information through self-attention before the next J-thought is
written. This ensures that the method does not preassign a topic or reasoning role to any
carrier. The single-carrier and no-inter-carrier-attention ablations in
Section~\ref{sec:ablation} evaluate these choices directly.

For input $x$, the prompt and neutral carrier embeddings are propagated through
the early Transformer blocks to a read layer $\ell_r$. Their carrier activations
at that point are denoted by $B_x^r\in\mathbb R^{M\times d}$. The prompt states
and $B_x^r$ are computed once and cached. Beginning every recurrent cycle from
this same prompt-conditioned baseline keeps the original problem available
while allowing differences between cycles to be attributable to the preceding
J-thought.

\subsection{Recurrent Read--Compute--Write Mechanism}
\label{sec:jcot-recurrence}

J-CoT places the read layer $\ell_r$ before a later write layer $\ell_w$.
Reading earlier gives the reconstructed state several Transformer blocks in
which to interact with the prompt and the other carriers. Writing later extracts
a state after that interaction has occurred. The two locations therefore define
a recurrent computation interval rather than a direct residual shortcut. Their
specific values are fixed for a backbone and selected on development data; they
are not adjusted separately for each benchmark.

Starting from $A_0=0$, cycle $t$ performs

\begin{equation}
\begin{aligned}
C_t^r
&=
B_x^r
+G_{\phi}^{r}(B_x^r,A_{t-1})
\odot\Psi_{\ell_r}(A_{t-1}),
\\
Z_t^w
&=
F_{\theta,x}^{\ell_r:\ell_w}(C_t^r),
\\
A_t
&=
\Phi_{\ell_w}(Z_t^w).
\end{aligned}
\label{eq:jcot-cycle}
\end{equation}

In the read operation, each row of the preceding J-thought is reconstructed
through the read-layer dictionary. The gate
$G_\phi^r\in[0,1]^{M\times d}$ regulates its contribution relative to the
prompt-conditioned carrier baseline. Addition is performed in the residual
stream because the reconstructed state is an input to computation, not the
complete carrier activation. The original prompt contribution is therefore not
replaced when the recurrent state is introduced.

The operator $F_{\theta,x}^{\ell_r:\ell_w}$ denotes the ordinary Transformer
blocks between the selected layers, conditioned on the cached prompt states.
Within this interval, the carriers use dense residual representations and may
interact with every permitted prompt and carrier position. No J-space projection
is applied inside these blocks. This preserves the backbone's native
computational capacity and confines the structural constraint to the recurrent
boundary.

At the write layer, $\Phi_{\ell_w}$ is applied independently to each carrier
activation in $Z_t^w$. The resulting rows form $A_t$. A coordinate persists
when the current hidden computation continues to express it at the write layer;
its coefficient changes when its strength changes, and it leaves the recurrent
state when it is no longer selected by the elastic-net solution. New coordinates
can enter in the same way. Retention and revision therefore arise from repeated
hidden computation followed by the same fixed extraction rule, rather than from
an explicit symbolic editing procedure.

J-CoT-Zero and J-CoT-Train share this recurrence. J-CoT-Zero fixes the neutral
carrier embeddings and uses unit read strength, so it adds no
interface-specific optimization to the reasoning-adapted backbone. J-CoT-Train
learns the carrier embeddings and the shared read gate while freezing the
Transformer and J-lens dictionaries. The learned component controls where the
model collects recurrent information and how strongly it is reintroduced; it
does not redefine the J-space coordinate system.

After recurrent execution, the final J-thought is converted back into carrier
states and used to condition autoregressive decoding. The stopping policy and
the exact answer-conditioning procedure are inference choices specified in
Section~\ref{sec:implementation}.

\section{Experiments}
\label{exp}

\subsection{Implementation Details}
\label{sec:implementation}

\paragraph{Architecture and recurrent execution.}
Our main experiments use \texttt{Qwen/Qwen3-8B-Base}~\citep{qwen2025qwen3},
with 36 Transformer layers and hidden dimension 4096. We append eight carrier positions after the
input prompt and use layer 12 for J-thought read-in and layer 28 for J-thought
extraction. Prompt activations and the neutral carrier baseline up to layer 12
are computed once and cached. At each cycle, the preceding J-thought is
reconstructed at layer 12, integrated through a shared read gate, propagated
through layers 13--28, and decomposed into the next J-thought using the
nonnegative elastic-net operator in Eq.~\ref{eq:j-coordinate-system}. We use
$\lambda_1=0.05$ and $\lambda_2=10^{-3}$.

\paragraph{Adaptive recurrence and answer conditioning.}
Rather than assigning the same recurrent depth to every input, we monitor the
normalized change in the J-thought,

\begin{equation}
r_t
=
\frac{\|A_t-A_{t-1}\|_F}
{\|A_t\|_F+\|A_{t-1}\|_F+\epsilon}.
\label{eq:jthought-residual}
\end{equation}

The numerator responds to changes in both active coordinates and their
coefficients, while normalization reduces sensitivity to the absolute scale of
the state. Recurrence terminates when $r_t<0.02$ for two consecutive cycles,
subject to a default maximum of eight cycles. Requiring two stable updates
avoids stopping on an isolated small transition. The scaling experiments use
maximum budgets of 4, 8, and 16 cycles; Appendix
gives the complete stopping procedure.

After stopping at cycle $\tau(x)$, the final J-thought is reconstructed at the
read layer and its carriers are propagated through the remaining Transformer
layers. The keys and values produced by these final carrier positions are
retained as fixed prefix key/value states during answer generation. Generated
tokens attend to the prefix through ordinary causal self-attention, while the
carrier positions themselves are excluded from the decoded output. This lets
the answer depend on the recurrent computation without requiring the J-thought
to be rendered as an intermediate rationale.

\paragraph{J-lens construction and interface training.}
The read- and write-layer J-lens dictionaries are constructed from randomized
low-rank estimates of the averaged downstream Jacobian on 1,000
pretraining-like sequences of 128 tokens. The dictionaries are computed once
per backbone and remain fixed. J-CoT-Zero uses the mean pretrained token
embedding as the fixed initialization of every carrier and applies unit-strength
J-thought read-in. J-CoT-Train optimizes the carrier embeddings and a shared
two-layer read gate with bottleneck width 256, while keeping the Transformer
and J-lens dictionaries frozen. Gradients through the elastic-net J-thought
extractor are computed by implicit differentiation of its converged active-set
solution. J-CoT-Train is optimized for 10,000 steps with
AdamW~\citep{loshchilov2019decoupled}, learning rate
$2\times10^{-4}$, weight decay 0.01, global batch size 128, and up to 16
unrolled recurrent cycles.

\paragraph{Evaluation and reproducibility.}
All methods begin from the same reasoning-adapted checkpoint and use the same
prompt templates, answer-processing rules, and decoding configuration. All
baselines that require additional training use identical Coconut-released
reasoning-step examples and train--development splits and follow Coconut's
published curriculum and checkpoint-selection protocol~\citep{hao2024coconut};
only their method-specific latent-state objectives differ. Sequential compute
is reported as measured pre-answer FLOPs, including prompt encoding, recurrent
Transformer blocks, and J-thought operations. Results are averaged over three
training seeds and reported with standard deviations and paired bootstrap
confidence intervals are used for benchmark-level comparisons. Dataset composition, duplicate filtering,
baseline-specific
optimization, software revisions, hardware, interface-spectrum construction,
and complete evaluation rules are provided in
Appendix. 

\subsection{Main Results}
\paragraph{Evaluation setting.}
For the main comparison, all methods are built on \texttt{Qwen3-8B-Base}.
Because the raw checkpoint is not instruction-tuned, we first construct a
shared reasoning-adapted model by supervised fine-tuning it on a common mixture
of mathematical, scientific, and programming problems with step-by-step
solutions. All evaluation instances, exact duplicates, and detected
near-duplicates are removed from this mixture. CoT and
PS+~\citep{wang2023plansolve} are applied directly to this shared checkpoint
without additional parameter updates. Coconut~\citep{hao2024coconut},
CODI~\citep{shen2025codi}, and SIM-Coconut~\citep{wei2026simcot} are
subsequently trained on the same Coconut-released training examples and
development splits using the published multi-stage curriculum.
Each baseline retains its original latent-state objective and reasoning
interface. J-CoT-Train uses the same interface-training examples and splits but
optimizes only the carrier embeddings and read gate using the schedule in
Section~\ref{sec:implementation}.
J-CoT-Zero uses the same shared reasoning-adapted checkpoint but introduces no
additional interface-specific optimization. This design ensures that all
methods share the same model initialization and domain adaptation while their
reasoning interfaces and training objectives follow the corresponding methods.

\paragraph{Compared methods.}
CoT~\citep{wei2022chain} is the standard explicit chain-of-thought baseline,
and PS+~\citep{wang2023plansolve} is the stronger linguistic-interface baseline
used in our comparison. Coconut~\citep{hao2024coconut},
CODI~\citep{shen2025codi}, and SIM-Coconut~\citep{wei2026simcot} are
latent-interface baselines that pass continuous internal states between
reasoning stages rather than appending a decoded rationale. J-CoT-Zero uses
fixed interface components, whereas J-CoT-Train learns the carrier embeddings
and read gate described in Section~\ref{sec:implementation}.

We evaluate mathematical and scientific reasoning on
GSM8K~\citep{cobbe2021gsm8k}, MATH-500~\citep{hendrycks2021math,lightman2023verify},
AIME~2024~\citep{maa2024aime}, and GPQA-Diamond~\citep{rein2024gpqa}, and code
reasoning on HumanEval+ and MBPP+~\citep{liu2023evalplus},
LiveCodeBench~\citep{jain2024livecodebench}, and
CRUXEval~\citep{gu2024cruxeval}. We report exact-match accuracy for mathematical
and scientific tasks, pass@1 under the official execution environments for
HumanEval+, MBPP+, and LiveCodeBench, and the macro-average of input- and
output-prediction accuracy for CRUXEval. The primary evaluation uses greedy
decoding with one solution per problem and excludes self-consistency, external
verifiers, retrieval, tool use, and execution-guided repair. Recurrent methods
are evaluated with a maximum of 8 reasoning cycles, while linguistic
methods receive a matched sequential-compute budget determined on the
development set. Hyperparameters are selected once using the aggregate
development set and are not tuned separately for individual benchmarks. We
report the mean over three random seeds.

\paragraph{Comparison with reasoning baselines.}
As shown in Table~\ref{tab:qwen3_8b_results}, J-CoT-Train outperforms both
linguistic and latent reasoning baselines across all eight benchmarks, achieves
the best result on every task, and improves the
overall average from 47.5 for the strongest competing latent method,
SIM-Coconut, to 50.2, corresponding to a gain of 2.7 points. The improvement
is consistent across domains: J-CoT-Train exceeds SIM-Coconut by 2.1--3.6
points on mathematical and scientific reasoning and by 2.6--3.2 points on
code reasoning. These results show that the benefit of J-CoT extends across
different task structures and answer formats, covering numerical reasoning,
scientific question answering, code generation, and program execution
reasoning.

J-CoT-Zero also matches or exceeds SIM-Coconut on every benchmark, increasing
the average score from 47.5 to 47.9 without J-CoT-specific training. It ties
SIM-Coconut on AIME~2024 and improves upon it by 0.3--0.6 points on the
remaining seven benchmarks. Compared with standard linguistic CoT,
J-CoT-Zero improves the average by 2.1 points, with per-task gains ranging
from 1.1 points on AIME~2024 to 2.8 points on MATH-500. This result supports
the hypothesis that the pretrained Transformer already contains a usable
J-space coordinate system from which recurrent J-thoughts can be constructed. The
read-gate calibration introduced by J-CoT-Train further raises the average
from 47.9 to 50.2 and improves performance on every benchmark, suggesting
additional value from learning how strongly the model should read its
preceding J-thought during each cycle.

The improvement remains substantial across both established and more
challenging benchmarks. Relative to SIM-Coconut, J-CoT-Train gains 3.6 points
on MATH-500, 3.2 points on CRUXEval, 2.8 points on HumanEval+ and
LiveCodeBench, 2.6 points on MBPP+, and 2.5 points on GPQA-Diamond. It also
improves AIME~2024 by 2.2 points and, even on the more saturated GSM8K
benchmark, provides a 2.1-point gain. This pattern is consistent with the
use of a recurrent J-thought in place of decoded tokens or a
dense hidden-state boundary.

\begin{table}[t]
\centering
\small
\caption{
Reasoning results under a shared \texttt{Qwen3-8B-Base} backbone. Values are
percentages averaged over three random seeds. Mathematical and scientific tasks
use accuracy; HumanEval+, MBPP+, and LiveCodeBench use pass@1; CRUXEval uses the
macro-average of input- and output-prediction accuracy. The final row is the
unweighted mean across the eight benchmarks. Higher is better.
}
\vspace{2pt}

\resizebox{\linewidth}{!}{
\begin{tabular}{ll|cc|ccc|cc}
\toprule
& &
\multicolumn{2}{c|}{\textbf{Linguistic CoT}} &
\multicolumn{3}{c|}{\textbf{Latent CoT}} &
\multicolumn{2}{c}{\textbf{Ours}} \\
\cmidrule(lr){3-4}
\cmidrule(lr){5-7}
\cmidrule(lr){8-9}

Category & Benchmark
& CoT
& PS+
& Coconut
& CODI
& SIM-Coconut
& J-CoT-Zero
& J-CoT-Train \\
\midrule

\multirow{3}{*}{\shortstack{Math}}
& GSM8K
& $82.6 \pm 0.5$
& $83.4 \pm 0.4$
& $81.8 \pm 0.7$
& $83.1 \pm 0.6$
& $84.0 \pm 0.5$
& $84.3 \pm 0.6$
& $\mathbf{86.1 \pm 0.4}$ \\

& MATH-500
& $48.2 \pm 0.9$
& $49.6 \pm 0.8$
& $47.1 \pm 1.1$
& $49.2 \pm 1.0$
& $50.4 \pm 0.9$
& $51.0 \pm 0.8$
& $\mathbf{54.0 \pm 0.7}$ \\

& AIME 2024
& $6.7 \pm 3.3$
& $7.8 \pm 3.8$
& $6.7 \pm 2.9$
& $6.7 \pm 3.4$
& $7.8 \pm 3.2$
& $7.8 \pm 3.5$
& $\mathbf{10.0 \pm 3.0}$ \\

\midrule

\shortstack{Science}
& GPQA-Diamond
& $34.2 \pm 1.3$
& $34.8 \pm 1.1$
& $33.7 \pm 1.5$
& $34.7 \pm 1.3$
& $35.5 \pm 1.0$
& $36.0 \pm 1.1$
& $\mathbf{38.0 \pm 0.9}$ \\

\midrule

\multirow{4}{*}{\shortstack{Coding}}
& HumanEval+
& $57.9 \pm 1.0$
& $59.1 \pm 0.9$
& $56.7 \pm 1.2$
& $58.7 \pm 1.0$
& $59.8 \pm 0.9$
& $60.2 \pm 0.8$
& $\mathbf{62.6 \pm 0.7}$ \\

& MBPP+
& $62.8 \pm 0.7$
& $63.8 \pm 0.6$
& $61.7 \pm 0.9$
& $63.4 \pm 0.8$
& $64.5 \pm 0.7$
& $65.0 \pm 0.6$
& $\mathbf{67.1 \pm 0.5}$ \\

& LiveCodeBench
& $19.4 \pm 0.8$
& $20.3 \pm 0.7$
& $18.9 \pm 1.0$
& $20.1 \pm 0.9$
& $21.0 \pm 0.8$
& $21.5 \pm 0.7$
& $\mathbf{23.8 \pm 0.6}$ \\

& CRUXEval
& $54.7 \pm 0.8$
& $55.8 \pm 0.7$
& $53.9 \pm 1.0$
& $55.5 \pm 0.8$
& $56.6 \pm 0.7$
& $57.2 \pm 0.7$
& $\mathbf{59.8 \pm 0.6}$ \\

\midrule

\multicolumn{2}{l|}{\textbf{Average}}
& $45.8 \pm 1.2$
& $46.8 \pm 1.1$
& $45.0 \pm 1.3$
& $46.5 \pm 1.2$
& $47.5 \pm 1.1$
& $47.9 \pm 1.1$
& $\mathbf{50.2 \pm 0.9}$ \\

\bottomrule
\end{tabular}}
\label{tab:qwen3_8b_results}
\vspace{6pt}
\end{table}

\subsection{Scaling with Model Capacity and Thinking Depth}
\label{sec:scalability}

We evaluate whether J-CoT scales jointly with backbone capacity and adaptive
test-time reasoning. Our study covers five dense pretrained models from 7B to
405B parameters: Qwen2.5-7B, Qwen2.5-14B, and
Qwen2.5-32B~\citep{yang2024qwen25}, and Llama-3.1-70B and
Llama-3.1-405B~\citep{dubey2024llama3}, denoted as \textbf{B}, \textbf{L}, \textbf{XL},
\textbf{XXL}, and \textbf{H}. For each backbone, we consider three adaptive
reasoning modes with maximum budgets of 4, 8, and 16 recurrent cycles, denoted
as \emph{Instant}, \emph{Medium}, and \emph{Heavy}.

\begin{figure}[t]
\centering
\begin{minipage}[t]{0.495\linewidth}
\centering
\includegraphics[width=\linewidth]{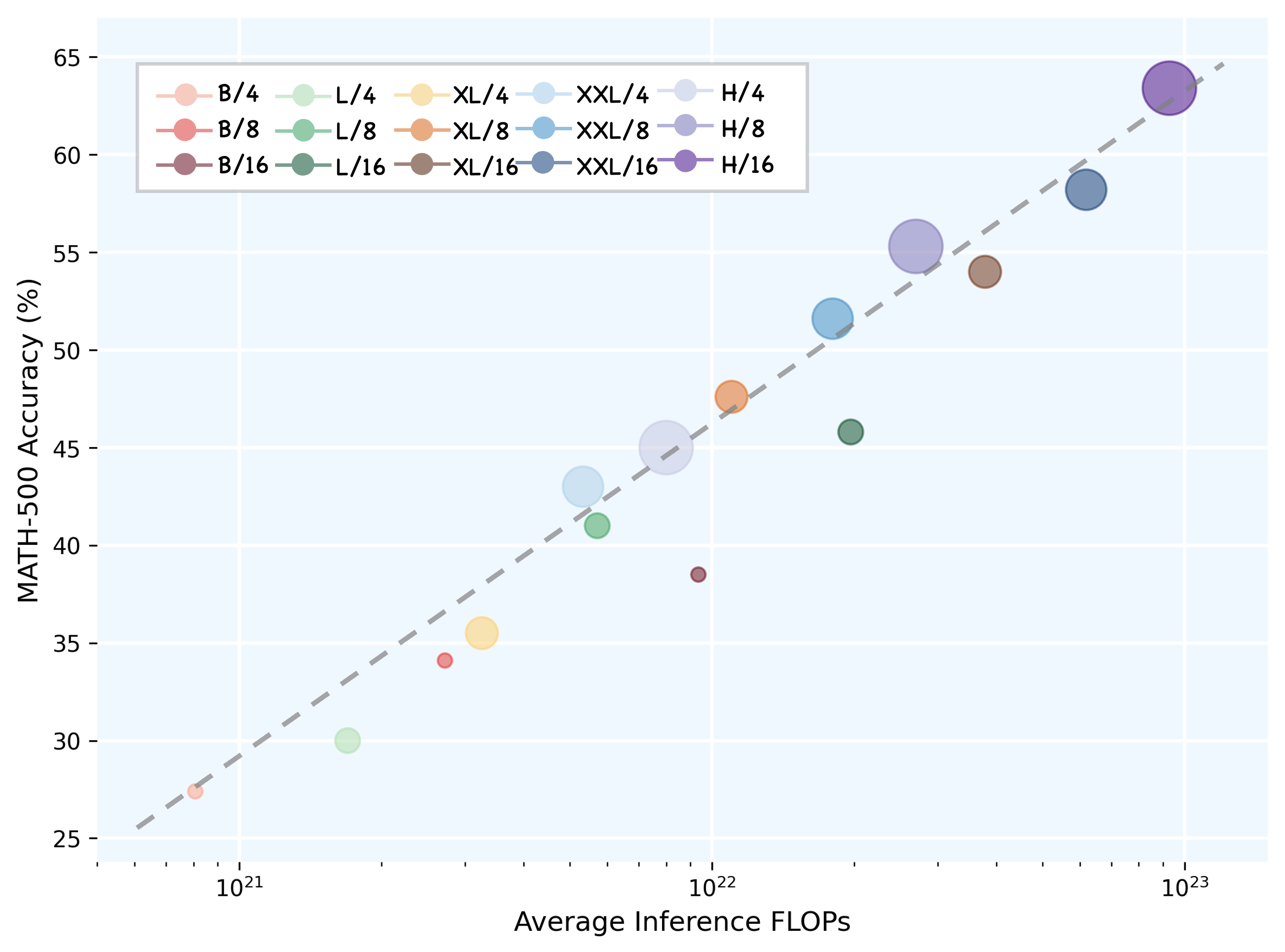}\\[-2pt]
\small (a) MATH-500
\end{minipage}\hfill
\begin{minipage}[t]{0.495\linewidth}
\centering
\includegraphics[width=\linewidth]{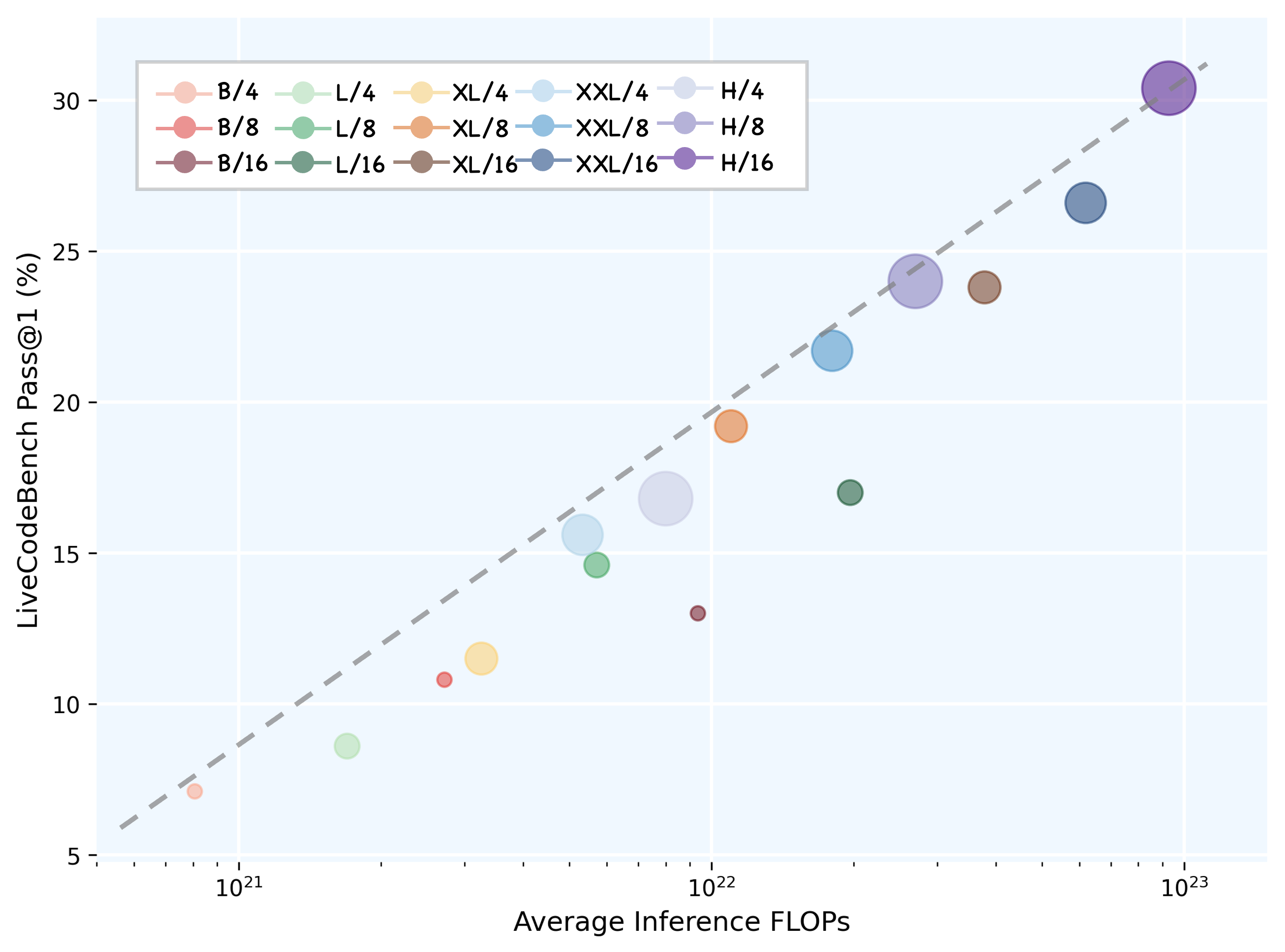}\\[-2pt]
\small (b) LiveCodeBench
\end{minipage}
\caption{J-CoT scaling with model capacity and recurrent depth. B, L, XL, XXL,
and H denote Qwen2.5-7B, Qwen2.5-14B, Qwen2.5-32B, Llama-3.1-70B, and
Llama-3.1-405B. The suffixes 4, 8, and 16 denote the maximum recurrent-cycle
budgets for Instant, Medium, and Heavy reasoning. Bubble area indicates model
scale, the horizontal axis reports measured average inference FLOPs, and the
dashed lines summarize the overall compute. Higher is
better.}
\label{fig:jcot-scaling}
\end{figure}

Figure~\ref{fig:jcot-scaling} reports all 15 configurations on MATH-500 and
LiveCodeBench. Each legend entry combines a backbone code with its maximum
cycle budget (e.g., H/16 denotes the 405B model with up to 16 cycles), and
bubble area indicates model scale. The horizontal axis reports measured average
inference FLOPs on a logarithmic scale, incorporating both backbone size and
realized reasoning depth.

J-CoT exhibits consistent scaling along both dimensions. On MATH-500, Heavy
reasoning improves over Instant from 27.4 to 38.5 for the B model
(+11.1 points) and from 45.0 to 63.4 for the H model (+18.4 points). On
LiveCodeBench, the corresponding improvements are from 7.1 to 13.0 for B
(+5.9 points) and from 16.8 to 30.4 for H (+13.6 points). At each cycle
budget, performance also increases with backbone capacity on both benchmarks.
The widening Heavy-over-Instant improvement from 7B to 405B indicates a
positive interaction between backbone capacity and recurrent reasoning depth,
rather than saturating at the scales studied, J-CoT obtains larger returns from
additional reasoning cycles on the largest model. This favorable scaling
behavior suggests that J-CoT may become still more effective as future language
models provide greater representational and computational capacity.

\FloatBarrier

\subsection{Reasoning Across the Interface Spectrum}
\label{sec:interface-spectrum}

J-CoT is designed as an intermediate recurrent interface between dense latent
reasoning and explicit linguistic CoT. To examine this position directly, we
construct a one-dimensional family of recurrent interfaces controlled by a
single parameter $\lambda\in[0,1]$. Let $h_t$ denote the dense latent state at
cycle $t$, let $j_t$ denote the J-thought state produced by the standard
J-CoT interface, and let $\ell_t$ denote the linguistic boundary state obtained
by decoding and re-encoding an explicit intermediate reasoning step. All three
states are normalized and represented in the same residual space.

We define
\begin{align}
w_{\mathrm{dense}}(\lambda)
    &= \max(1-2\lambda,0), \\
w_{\mathrm{J}}(\lambda)
    &= 1-\lvert 2\lambda-1\rvert, \\
w_{\mathrm{lang}}(\lambda)
    &= \max(2\lambda-1,0),
\end{align}
and transmit
\begin{equation}
b_t^{(\lambda)}
=
w_{\mathrm{dense}}(\lambda)\bar h_t
+
w_{\mathrm{J}}(\lambda)\bar j_t
+
w_{\mathrm{lang}}(\lambda)\bar \ell_t.
\label{eq:one-dimensional-interface}
\end{equation}
At $\lambda=0$, the recurrent boundary carries the dense hidden state. At
$\lambda=0.5$, it carries the J-thought state. At $\lambda=1$, it
carries an explicitly linguistic state. Intermediate values smoothly connect
the dense endpoint to J-CoT and J-CoT to the linguistic endpoint. 

\begin{figure}[t]
\centering
\includegraphics[width=\linewidth]{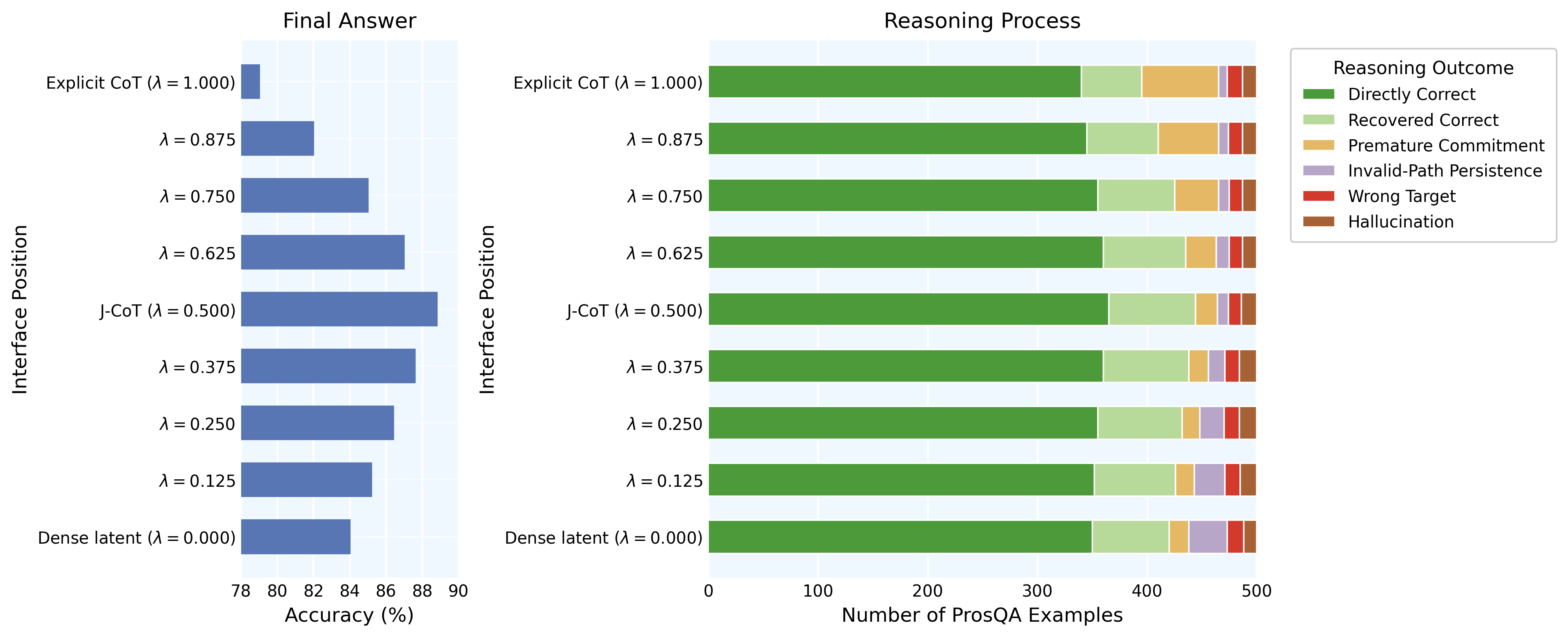}
\vspace{-8pt}
\caption{
Reasoning behavior across the recurrent-interface spectrum. We continuously
interpolate the recurrent interface between dense latent recurrence
($\lambda=0$), J-thought recurrence ($\lambda=0.5$), and explicit
linguistic recurrence ($\lambda=1$). The left panel reports final-answer
accuracy, while the right panel decomposes each prediction into mutually
exclusive reasoning outcomes. Performance peaks at the J-CoT configuration in
the evaluated sweep.
}
\label{fig:interface-spectrum}
\vspace{-10pt}
\end{figure}

We evaluate nine evenly spaced values of $\lambda$ on
ProsQA~\citep{hao2024coconut} while varying the branching factor and number of
distractor paths. Figure~\ref{fig:interface-spectrum}
shows a clear non-monotonic trend across the recurrent-interface spectrum.
Performance rises from 79.0\% at the explicit-language endpoint
($\lambda=1$) to 84.0\% at the dense latent endpoint ($\lambda=0$), and reaches
its maximum of 88.8\% at the J-CoT configuration, $\lambda=0.5$.

The reasoning-process decomposition is consistent with this pattern. Moving
away from the linguistic endpoint reduces premature commitment and increases
the number of cases in which the model recovers after initially selecting an
incorrect branch. Dense latent recurrence preserves these alternatives more
effectively than explicit CoT, but it also exhibits higher invalid-path
persistence, indicating that irrelevant branches remain active across recurrent
cycles. As $\lambda$ approaches the J-CoT region, recovery remains high while
invalid-path persistence is progressively reduced.

Among the evaluated interpolation points, the J-CoT configuration at
$\lambda=0.5$ attains the highest final-answer accuracy. The experiment
therefore locates the best observed operating point within this particular
interface sweep without assigning semantic properties to the transmitted state.

\FloatBarrier

\subsection{Ablation Study}
\label{sec:ablation}

We conduct ablations to validate the main design choices of J-CoT, summarized
in Table~\ref{tab:ablations}. All
experiments use Qwen3-8B-Base with the
same training data, reasoning budget, and decoding configuration. We report
average accuracy across MATH-500, AIME 2024, HumanEval+, and ProsQA.

\paragraph{Recurrent interface.}
Removing J-thought reading yields an average accuracy of 49.5, compared with
53.9 for the full configuration. Replacing the pretrained J-space with a
learned latent state yields 50.9, and replacing canonical J-space transport
with a learned transport adapter yields 51.5.

\paragraph{J-thought structure.}
Collapsing the J-thought into a single carrier reduces accuracy to 51.8,
compared with 53.9 for the full model. Removing interaction among carriers
reaches 52.6. A fixed-size J-thought with six coordinates per carrier obtains
53.0, while the adaptive configuration obtains 53.9.

The full configuration combines a persistent J-space interface with an adaptive
multi-carrier J-thought.

\begin{table}[t]
\centering
\caption{Ablations on Qwen3-8B-Base. The left panel tests the recurrent
interface, and the right panel tests J-thought structure. Values are average
accuracy (\%) across MATH-500, AIME 2024, HumanEval+, and ProsQA; higher is
better.}
\label{tab:ablations}
\vspace{2pt}
\begin{minipage}[t]{0.485\linewidth}
\centering
\scriptsize
\textbf{(a) Recurrent interface}\\[3pt]
\begin{tabularx}{\linewidth}{@{}>{\raggedright\arraybackslash}X r@{}}
\toprule
\textbf{Ablation variant} & \textbf{Avg. acc. $\uparrow$} \\
\midrule
\textit{w/o J-Thought Reading} & 49.5 \\
\textit{Learned Latent State} & 50.9 \\
\textit{Learned Transport Adapter} & 51.5 \\
\midrule
\textbf{J-CoT (full)} & \textbf{53.9} \\
\bottomrule
\end{tabularx}
\end{minipage}
\hfill
\begin{minipage}[t]{0.485\linewidth}
\centering
\scriptsize
\textbf{(b) J-thought structure}\\[3pt]
\begin{tabularx}{\linewidth}{@{}>{\raggedright\arraybackslash}X r@{}}
\toprule
\textbf{Ablation variant} & \textbf{Avg. acc. $\uparrow$} \\
\midrule
\textit{Single J-Thought Carrier} & 51.8 \\
\textit{No Inter-Carrier Attention} & 52.6 \\
\textit{Fixed-Size J-Thought ($k=6$)} & 53.0 \\
\midrule
\textbf{J-CoT (full)} & \textbf{53.9} \\
\bottomrule
\end{tabularx}
\end{minipage}
\end{table}

\FloatBarrier

% \section{Discussion and Limitations}

% J-CoT organizes recurrence around a J-thought rather
% than a decoded rationale or an unrestricted hidden activation. The recurrent
% boundary is defined by this vocabulary-indexed state and the associated read and write
% operators. The ablations evaluate J-thought reading, canonical transport, and
% multi-carrier structure.

% Several limitations remain. First, vocabulary indices should not be assumed to
% correspond to complete human concepts: tokenizer fragments and distributed
% representations may contribute to a coordinate. Second, repeated sparse coding
% and J-lens calibration introduce
% computational overhead beyond the Transformer forward passes. Third, the scaling
% study spans two model families, so differences between the largest and smallest
% backbones reflect both capacity and family-specific training or architecture.
% Finally, the present evaluation focuses on a selected set of mathematical,
% scientific, coding, and structured path-reasoning tasks. Broader evaluations,
% uncertainty analysis, and direct latency and memory measurements are important
% directions for future work.

\section{Conclusion}

We introduced J-CoT, a recurrent reasoning framework that reads and writes a
J-thought in the model's vocabulary-indexed J-space. The model retains
unrestricted hidden computation within each cycle while using that coefficient
state at recurrent boundaries. Across the evaluated tasks,
J-CoT-Zero matches or exceeds the strongest evaluated latent baseline, while
J-CoT-Train improves further. The interface and J-thought ablations evaluate
the main design choices. These
results motivate further study of model-native recurrent interfaces
as an alternative to both fully verbalized and fully dense intermediate states.

\bibliographystyle{iclr2026_conference}
\bibliography{iclr2026_conference}

\end{document}